\documentclass{article}

\usepackage[preprint]{neurips_2026}

\usepackage[utf8]{inputenc} 
\usepackage[T1]{fontenc} 
\usepackage{hyperref} 
\usepackage{url} 
\usepackage{booktabs} 
\usepackage{amsfonts} 
\usepackage{nicefrac} 
\usepackage{microtype} 
\usepackage{xcolor} 
\usepackage{graphicx}
\usepackage{amsmath}
\usepackage{booktabs}
\usepackage{multirow}

\usepackage{xcolor} 
\usepackage{colortbl}
\definecolor{mygray}{gray}{0.95}
\usepackage{xurl}
\usepackage{url}

\usepackage[most]{tcolorbox}
\usepackage{xcolor}
\usepackage{enumitem}

\definecolor{traceText}{HTML}{263238}
\definecolor{traceBorder}{HTML}{CBD5D8}
\definecolor{traceTitle}{HTML}{37474F}

\definecolor{userBack}{HTML}{F3F7FA}
\definecolor{userFrame}{HTML}{AFC3CF}
\definecolor{userTitle}{HTML}{E5EEF3}

\definecolor{modelBack}{HTML}{F7F7F5}
\definecolor{modelFrame}{HTML}{C8C6BE}
\definecolor{modelTitle}{HTML}{ECEAE4}

\definecolor{toolCallBack}{HTML}{FAF4EC}
\definecolor{toolCallFrame}{HTML}{D8C2A8}
\definecolor{toolCallTitle}{HTML}{F0E3D2}

\definecolor{toolResBack}{HTML}{F1F7F4}
\definecolor{toolResFrame}{HTML}{B8CFC2}
\definecolor{toolResTitle}{HTML}{E1EEE7}

\definecolor{finalBack}{HTML}{F3F8F0}
\definecolor{finalFrame}{HTML}{BACDAE}
\definecolor{finalTitle}{HTML}{E4EEDD}

\definecolor{tagPurple}{HTML}{6D5A7A}

\tcbset{ agenttracebase/.style={ enhanced, breakable, parbox=false, boxrule=0.55pt, arc=1.4mm, left=1.2mm, right=1.2mm, top=1mm, bottom=1mm, before skip=0.42em, after skip=0.42em, fonttitle=\bfseries\small, coltitle=traceTitle, coltext=traceText, borderline west={1.2pt}{0pt}{traceBorder} } }

\newtcolorbox{userbox}[1]{ agenttracebase, colback=userBack, colframe=userFrame, colbacktitle=userTitle, title={#1}, borderline west={1.5pt}{0pt}{userFrame} }

\newtcolorbox{modelbox}[1]{ agenttracebase, colback=modelBack, colframe=modelFrame, colbacktitle=modelTitle, title={#1}, borderline west={1.5pt}{0pt}{modelFrame} }

\newtcolorbox{toolcallbox}[1]{ agenttracebase, colback=toolCallBack, colframe=toolCallFrame, colbacktitle=toolCallTitle, title={#1}, borderline west={1.5pt}{0pt}{toolCallFrame} }

\newtcolorbox{toolresbox}[1]{ agenttracebase, colback=toolResBack, colframe=toolResFrame, colbacktitle=toolResTitle, title={#1}, borderline west={1.5pt}{0pt}{toolResFrame} }

\newtcolorbox{finalbox}[1]{ agenttracebase, colback=finalBack, colframe=finalFrame, colbacktitle=finalTitle, title={#1}, borderline west={1.5pt}{0pt}{finalFrame} }

\title{
OpenClaw-Skill: Collective Skill Tree Search for Agentic Large Language Models
}

\author{%
\textbf{Tianyi Lin$^1$\thanks{Equal contribution.},
Chuanyu Sun$^1$\footnotemark[1],
Jingyi Zhang$^2$,
Changxu Wei$^1$,
Huanjin Yao$^3$,
Shunyu Liu$^2$,}\\
\textbf{Xikun Zhang$^4$,
Liu Liu$^5$,
Jiaxing Huang$^1$\thanks{Corresponding author.}}\\
\normalfont $^1$The Hong Kong Polytechnic University\\
\normalfont $^2$Nanyang Technological University\\
\normalfont $^3$Tsinghua University\\
\normalfont $^4$Royal Melbourne Institute of Technology\\
\normalfont $^5$Beijing University of Aeronautics and Astronautics
}

\begin{document}
  \maketitle

  \begin{abstract}
    Equipping Large Language Model (LLM) agents with effective skills is crucial
    for solving complex tasks in real-world systems like OpenClaw.
    In this work, we aim to develop a framework that automatically constructs such
    reusable skills to enhance LLMs in tool use, multi-step reasoning, and dynamic
    environment interaction.
    To this end, we propose Collective Skill Tree Search (CSTS), a novel tree-search-based
    skill construction framework that constructs structured, diverse and generalizable
    \textit{tree of skills}.
    The core idea of CSTS is to leverage collective intelligence to jointly search,
    identify and compose effective skills via two iterative phases: Collective Skill
    Node Generation (CSN-Gen) and Collective Skill Node Assessment (CSN-Assess).
    CSN-Gen exploits collective knowledge from multiple models to explore
    diverse candidate skills for each subtask, enabling comprehensive skill exploration.
    CSN-Assess employs multiple models as judges to evaluate and select skill
    nodes with two scoring mechanisms: (1) collective quality scoring that aggregates
    independent evaluations to produce a robust estimate of skill effectiveness,
    and (2) collective transferability scoring that explicitly verifies whether
    a skill generalizes well across different models.
    With CSTS, we construct a set of comprehensive \textit{tree of skills} along
    with skill-augmented training data, enabling models to effectively learn and
    utilize skills.
    Besides, we introduce Collective Skill Reinforcement Learning, which
    actively selects multiple relevant skills from the tree to broaden solution-space
    exploration, avoid being trapped by a single skill and its resulting homogeneous
    or suboptimal solutions.
    As a result, our trained model, OpenClaw-Skill, exhibits outstanding agentic
    capabilities in long-horizon planning, tool use and generalization over
    challenging benchmarks.
  \end{abstract}

  \section{Introduction}
  Large Language Model (LLM) agents have recently demonstrated strong potential
  for solving complex real-world tasks through natural language instructions~\cite{gpt5,claude_sonnet_4,
  gemini2.5, vicuna}, particularly in interactive environments such as OpenClaw~\cite{openclaw},
  where LLM agents~\cite{treesearchagents,react,toolformer} are required to
  coordinate files, tools, web pages, execution feedback, and intermediate
  artifacts over multiple steps. To further improve the reliability and generalization
  of such OpenClaw-like systems, recent studies have introduced the concept of ``skills''~\cite{skillrl,skillcraft,skillsbench},
  which encapsulate compact and reusable procedural strategies for accomplishing
  recurring subtasks, such as tool use, verification, and error recovery.

  However, current skills~\cite{anthropic2025agentSkills} are largely
  handcrafted and require substantial manual design and maintenance, making large-scale
  skill construction costly, time-consuming, and labor-intensive, thereby
  significantly limiting scalability.

  To address this issue, recent studies~\cite{skillrl,evoskill,coevoskills} have
  explored automatic skill construction and maintenance. For example, by distilling
  LLM agent execution experiences and raw trajectories into reusable procedural knowledge,
  SkillRL~\cite{xia2026skillrl} learns a skill bank from past interactions to
  facilitate policy improvement while Trace2Skill~\cite{trace2skill}
  consolidates trajectory-local lessons and common patterns into agent skills. In
  addition, CoEvoSkills~\cite{coevoskills} enables skill refinement by
  integrating skill generation with surrogate verification, allowing skills to be
  iteratively evaluated and revised according to execution feedback.

  \begin{figure}[t]
    \centering
    \includegraphics[width=\textwidth]{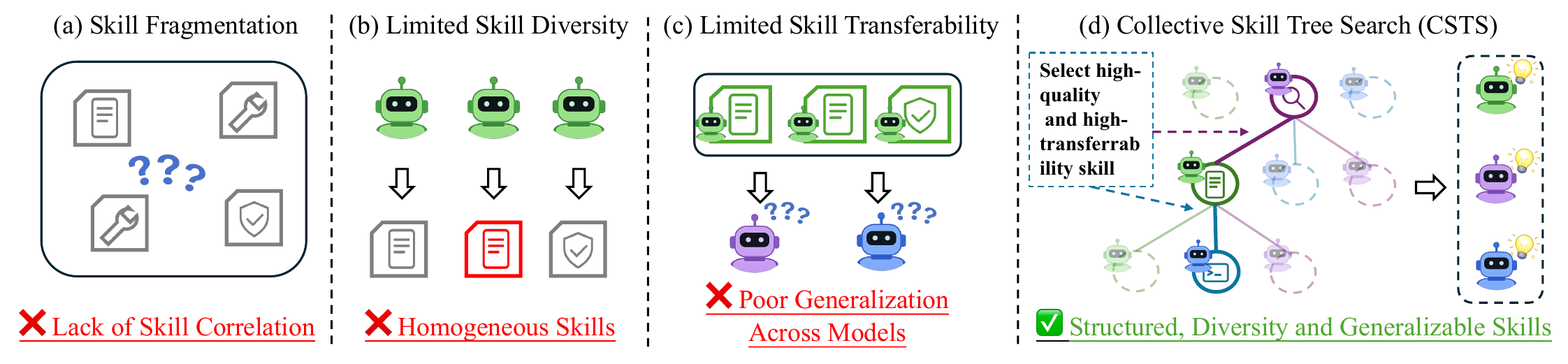}
    \caption{ \textbf{Motivation.} Current skill construction paradigms
    generally face several limitations: \textit{(a) Skill Fragmentation}, capturing
    merely local procedures for isolated subtasks; \textit{(b) Limited Diversity},
    suffering from the inherent biases of a single model; and \textit{(c) Poor
    Transferability}, exhibiting clear performance drops across different LLM backbones.
    To tackle these challenges, we propose CSTS, a novel tree-search-based skill
    construction framework that constructs structured, diverse, and
    generalizable \textit{tree of skills}, empowering LLMs in solving
    sophisticated tasks in real-world systems.
    }
    \label{fig:fig1}
  \end{figure}

  Despite these advances, most current paradigms for skill construction and
  maintenance still face several limitations, particularly for long-horizon tasks
  in real-world interactive environments such as OpenClaw, as illustrated in
  Figure~\ref{fig:fig1}: (1) \textbf{Skill Fragmentation}: Current methods often
  produce fragmented and unstructured skills that merely capture local procedures
  for individual subtasks, but lack explicit mechanisms to orchestrate skill
  sequences, handle long-term dependencies, and verify execution across multiple
  steps; (2) \textbf{Limited Skill Diversity}: existing approaches typically construct
  the skill from a narrow set of trajectories generated by a single model. As a
  result, the constructed skills are inherently biased toward the problem-solving
  preferences of that model, limiting their coverage across diverse task types
  and reasoning strategies. (3) \textbf{Limited Skill Transferability}: Skills acquired
  through current methods often exhibit limited generalization ability. In
  particular, their performance tends to degrade clearly when transferred to different
  backbone LLMs.

  To tackle these challenges, we propose Collective Skill Tree Search (CSTS), a
  new tree-search-based skill construction framework that constructs structured,
  diverse and generalizable skills for LLM agents, enhancing their capability in
  solving sophisticated real-world tasks. Specifically, CSTS works by leveraging
  collective intelligence to jointly search, identify and compose effective skills
  iteratively towards cohesive and reliable ``tree of skills''. CSTS operates through
  two iterative phases: \textit{Collective Skill Node Generation} (CSN-Gen) and \textit{Collective
  Skill Node Assessment} (CSN-Assess). In skill node generation phase, CSN-Gen
  exploits collective knowledge from multiple models to explore diverse
  candidate skills for the current subtask, enabling rich and comprehensive skill
  discovery. In skill node assessment phase, CSN-Assess employs multiple models as
  judges to evaluate and select skill nodes with strong effectiveness and transferability
  through two scoring mechanisms. The first is \textit{collective skill quality
  scoring}, where multiple judges independently assess the skill quality and
  aggregate their evaluations into a robust score. The second is \textit{collective
  skill transferability scoring}, which works by measuring whether a skill synthesized
  by one model can effectively benefit other models, thereby explicitly
  encouraging the selection of highly generalizable skills.

  In this way, our proposed CSTS enables structured, diverse, and transferable
  skill construction: (1) ``Tree of skills'' mitigates skill fragmentation. By modeling
  skills as nodes and subtask-solving procedures as paths, CSTS transforms
  isolated local skills into structured, dependency-aware skill hierarchies,
  enabling LLM agents to effectively compose and orchestrate skill sequences
  across complex multi-stage tasks. (2) The design of collective skill node generation
  enhances skill diversity. By aggregating candidate skills from multiple
  heterogeneous models, CSN-Gen explicitly alleviates single-model bias and enriches
  the skill space with diverse and complementary behavioral patterns, thereby
  reducing overfitting to specific reasoning styles. (3) The design of collective
  skill transferability scoring improves skill transferability. By requiring
  skills synthesized by one model to effectively benefit other models, this mechanism
  explicitly encourages capturing skills that generalize well across different backbone
  LLMs, leading to generalizable performance.

  Based on our proposed CSTS, we construct a set of comprehensive \textit{tree
  of skills} together with skill-augmented training data for complex tasks under
  OpenClaw-style systems. By using these skill-augmented training data, we first
  train our model via Supervised Fine-Tuning (SFT). In addition, we propose Collective
  Skill Reinforcement Learning (CSRL) to further optimize the model and obtain our
  final model, termed OpenClaw-Skill. Specifically, for each task, our CSRL actively
  selects multiple relevant skills from the tree of skills to substantially broaden
  the exploration of the solution space, thereby preventing the model from being
  trapped in a single skill and its resulted homogeneous or suboptimal solutions.
  As a result, the model is allowed to adaptively fit the most appropriate skill
  among multiple candidate skills, significantly improving the flexibility,
  robustness, and effectiveness of skill reinforcement learning.

  The main contributions of this work are summarized as follows. First, we
  introduce Tree Search into skill construction for LLM agents and propose
  Collective Skill Tree Search (CSTS), which leverages collective intelligence to
  jointly search, identify and compose effective skills iteratively towards
  cohesive and reliable ``tree of skills''.
  Second, we construct a set of comprehensive tree of skills and skill-augmented
  training data for complex tasks in OpenClaw-style systems, enabling models to learn
  procedural knowledge from collective agent experiences.
  Third, we introduce Collective Skill Reinforcement Learning (CSRL), which
  further optimize capabilities of the model by actively selecting multiple
  skills to broaden the exploration of the solution space, thereby preventing the
  model from being trapped in a single skill and its induced homogeneous and low-quality
  solutions.
  Fourth, we develop OpenClaw-Skill, a series of models with outstanding
  capabilities in long-horizon planning, tool use, error recovery and cross-task
  generalization, demonstrating superior performance on multiple challenging benchmarks.

  \section{Related Works}

  \subsection{Large Language Model Agents}
  Recent advances have transformed large language models~\cite{qwen2024qwen25,qwen2025qwen3,yang2024qwen2,deepseek2025r1,deepseek2024v3,deepseek2024v2,deepseek2024llm,dai2024deepseekmoe,ouyang2022instructgpt}
  from static text generators into interactive agents capable of interleaving reasoning,
  action, and environment feedback ~\cite{plaat2025agentic,luo2025llmagent,chen2024llmmas,wang2023autonomous,zhang2025agenticrl}.
  ReAct~\cite{react} first established a simple yet influential reasoning–acting
  paradigm for sequential decision-making and tool use, while WebVoyager~\cite{webvoyager}
  and OSWorld~\cite{osworld} pushed LLM agents toward more realistic web and
  computer-use tasks. This shift has driven the emergence of agent harnesses that
  expose models to filesystems, browsers, and desktop applications. OpenClaw~\cite{openclaw}
  firstly exemplifies a move beyond toy environments toward real-world
  persistent runtimes with messaging interfaces, sessions, tools, and structured
  workspace state. Building on this paradigm, DeerFlow~\cite{deerflow} and Hermes~\cite{hermes}
  further extend OpenClaw by enabling richer access to memory, sandboxed, and communication.
  ClawGym~\cite{clawgym} and OpenClaw-RL~\cite{openclawrl} take initial steps
  toward studying data generation and model training in such Claw-style
  environments. Nevertheless, how to automatically construct effective skills to
  guide LLM agents to accomplish complex tasks in real-world systems remains a
  significant open challenge.

  \subsection{Skills for Large Language Model Agents}
  Anthropic first introduced the concept of skills~\cite{anthropic2025agentSkills}
  as reusable procedural knowledge that encapsulates instructions and tool-usage
  patterns, enabling agents to dynamically extend their capabilities at inference
  time. Recently, a growing body of work has begun to explore the evaluation,
  generation, and utilization of LLM skills~\cite{zhang2026skillflow,wang2023voyager,jiang2026sokskills,xu2026agentskills,yang2026autoskill,coevoskills,trace2skill,evoskill,skillrl,skillcraft,xskill,skill0}.
  For example, SkillRL~\cite{skillrl} constructs a SkillBank and learns skill invocation
  policies via reinforcement learning. Trace2Skill~\cite{trace2skill}
  consolidates trajectory-local lessons into skill repositories.
  Despite recent progress, current skill construction paradigms still suffer
  from several limitations, especially for long-horizon tasks in real-world environments
  such as OpenClaw, including skill fragmentation, limited skill diversity, and
  weak skill transferability. Different from previous approaches, our CSTS
  introduces a novel tree-search-based skill construction framework that builds
  structured, diverse, and generalizable \textit{trees of skills}, empowering
  LLM agents to solve sophisticated tasks in real-world systems.

  \subsection{Tree Search}
  Tree search has emerged as an important paradigm for improving decision-making
  and LLM reasoning~\cite{treesearchagents,treeofthought,mulberry,lats,silver2017mastering,ye2021mastering}.
  A representative example is AlphaGo~\cite{silver2017mastering}, which combines
  neural network priors with tree search to enable strategic planning and
  decision-making, achieving remarkable performance in complex board and video game
  environments~\cite{silver2017mastering,ye2021mastering}. Tree Search for
  Language Model Agents showed that explicit search over action trajectories can
  substantially improve success rates in realistic web environments
  \cite{treesearchagents}.
  Inspired by these advances, we introduces the concept of ``tree search'' into automatic
  skill construction, and propose CSTS, a novel framework that decomposes complex
  tasks into subtasks and leverages collective intelligence to jointly search,
  identify and compose effective skills iteratively towards cohesive and
  reliable ``tree of skills''.

  \section{Method}
  We first present \textbf{Collective Skill Tree Search} (CSTS) that constructs structured,
  diverse, and transferable tree of skills via collective intelligence, and
  describe training data construction with CSTS and model training. We then introduce
  \textbf{Collective Skill Reinforcement Learning} (CSRL) that further optimizes
  the model by comparing trajectories conditioned on different skills under the
  same subtask.

  \subsection{Collective Skill Tree Search}
  The core idea of CSTS is to utilize collective intelligence to jointly search,
  identify and compose effective skill iteratively towards cohesive and reliable
  ``tree of skills''. Specifically, CSTS first conducts (1) \textit{Complex Task
  Decomposition}, and then operates via two iterative phases, including (2)
  \textit{Collective Skill Node Generation (CSN-Gen)} and (3) \textit{Collective
  Skill Node Assessment (CSN-Assess)}, which together organize skills into a tree,
  where each layer corresponds to a subtask and each node denotes a candidate
  skill. A path through the tree represents a compositional skill path that specifies
  how local skills are selected and ordered across subtasks.

  \textbf{Complex Task Decomposition.} Given a complex task $T$, CSTS first decomposes
  it into an ordered sequence of subtasks:
  \[
    T \rightarrow (t_{1},t_{2},\ldots,t_{M}).
  \]
  This decomposition identifies the main procedural stages, such as locating files,
  inspecting configurations, constructing commands, executing tools, diagnosing failures,
  and verifying outputs. It defines the skill-tree depth: the $m$-th layer
  corresponds to subtask $t_{m}$. CSTS then performs skill generation and assessment
  for each subtask layer.

  \begin{figure}[t]
    \centering
    \includegraphics[width=\textwidth]{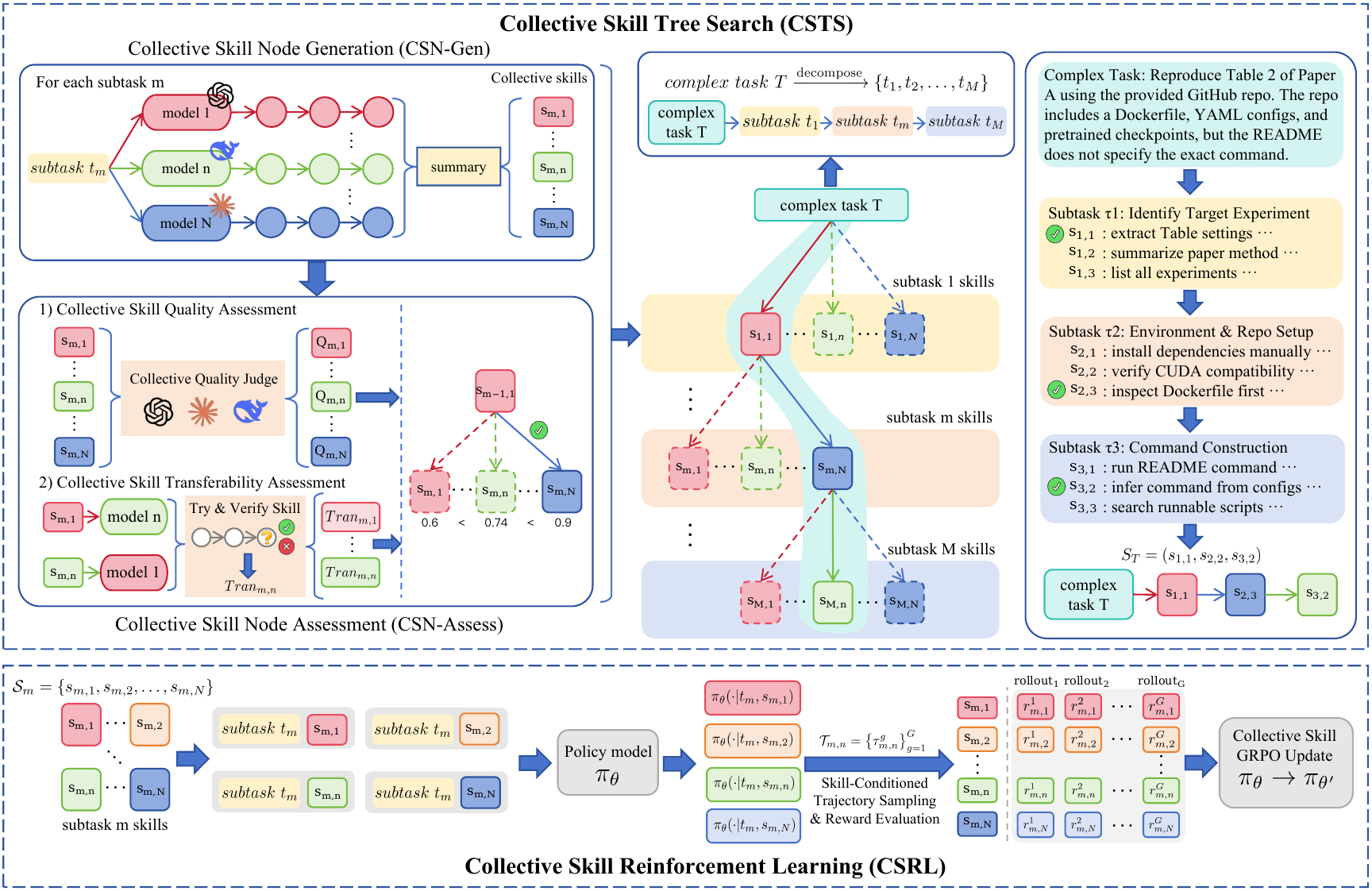}
    \caption{ \textbf{Overview of our OpenClaw-Skill with CSTS and CSRL.} Given
    a complex agentic task, CSTS first decomposes it into subtasks and iteratively
    constructs skill nodes for each subtask. For each subtask, CSN-Gen summarizes
    diverse trajectories from multiple heterogeneous agents into candidate collective
    skill nodes. CSN-Assess then evaluates and selects these nodes via
    collective skill quality assessment and collective skill transferability
    assessment. The selected skill nodes form a tree-structured search space,
    where each path represents a compositional skill plan capturing stage-wise
    dependencies across subtasks. The bottom panel shows CSRL, where the model
    generates skill-conditioned trajectory groups and is optimized with a
    Collective Skill GRPO Update using relative advantages across skills under
    the same subtask. }
    \label{fig:fig2}
  \end{figure}

  \textbf{Collective Skill Node Generation (CSN-Gen).} For each subtask $t_{m}$,
  CSTS collectively generates multiple candidate skill nodes from trajectories
  produced by a group of models. Let
  \[
    \mathcal{M}=\{M_{1},M_{2},\ldots,M_{N}\}
  \]
  denote the set of participating models. Each model $M_{n}$ attempts to solve
  the same subtask $t_{m}$ and produces an execution trajectory:
  \[
    \tau_{m,n}= \pi_{\theta_n}(\cdot \mid t_{m}) = \left( t_{m}, \{(\psi_{m,n}^{\ell}
    ,a_{m,n}^{\ell},o_{m,n}^{\ell})\}_{\ell=1}^{L_{m,n}}, r_{m,n}\right),
  \]
  where $\psi_{m,n}^{\ell}$, $a_{m,n}^{\ell}$, and $o_{m,n}^{\ell}$ denote the intermediate
  reasoning state, agent action, and observation or execution feedback at step $\ell$,
  respectively. The action $a_{m,n}^{\ell}$ may correspond to a tool call, file
  operation, code execution, or textual response, depending on the interaction context.
  The scalar $r_{m,n}$ denotes the final execution outcome or correctness signal
  of the trajectory.

  Collective skill node generation mitigates single-model bias by exposing skill
  construction to diverse procedural evidence for the same subtask. Different models
  may explore different solution routes, encounter different failure modes, and reveal
  different verification opportunities. This diversity is important for
  constructing candidate skills that cover a broader range of planning, execution,
  diagnosis, and recovery patterns.

  To efficiently obtain diverse candidate skills, CSTS uses a shared skill
  synthesizer $\Phi_{\mathrm{skill}}$ to summarize each trajectory into a
  candidate skill node:
  \[
    s_{m,n}= \Phi_{\mathrm{skill}}(t_{m},\tau_{m,n}).
  \]
  The resulting candidate skill set for subtask $t_{m}$ is
  \[
    \mathcal{S}_{m}= \{s_{m,1},s_{m,2},\ldots,s_{m,N}\}.
  \]
  Each skill $s_{m,n}$ describes a reusable procedure for solving $t_{m}$,
  including its applicable context, required inputs, recommended actions, expected
  outputs, verification criteria, and recovery strategies. Since trajectory
  generation is performed in parallel across models and skill synthesis is
  applied uniformly to each trajectory, CSN-Gen improves both the efficiency and
  diversity of candidate skill construction.

  \textbf{Collective Skill Node Assessment (CSN-Assess).} After generating
  candidate skills for each subtask, CSTS evaluates each skill node from two complementary
  perspectives: collective skill quality and collective skill transferability. For
  collective skill quality assessment, multiple judge models independently
  evaluate whether a skill is clear, executable, complete, and relevant to the target
  subtask. Given the judge score $q_{m,n}^{j}$ assigned by judge $j$, the collective
  quality score of skill $s_{m,n}$ is defined as
  \[
    Q_{m,n}= \frac{1}{J}\sum_{j=1}^{J}q_{m,n}^{j}.
  \]

  For collective skill transferability assessment, CSTS further measures whether
  a skill synthesized from one model can benefit other models. Specifically, skill
  $s_{m,n}$, distilled from the trajectory of model $M_{n}$ on subtask $t_{m}$, is
  shared with the remaining models $\{M_{k}\}_{k\neq n}$. Each model $M_{k}$
  then uses $s_{m,n}$ as additional context to solve the same subtask $t_{m}$, producing
  a skill-conditioned rollout:
  \[
    \tilde{\tau}_{m,n}^{k}\sim \pi_{\theta_k}(\cdot \mid t_{m}, s_{m,n}), \qquad
    k\neq n.
  \]
  Let $r_{m,n}^{k}$ denote the verification score of the rollout
  $\tilde{\tau}_{m,n}^{k}$. The transferability score is computed over the $N-1$
  models that did not produce the original skill:
  \[
    \mathrm{Tran}_{m,n}= \frac{1}{N-1}\sum_{\substack{k=1 \\ k\neq n}}^{N}r_{m,n}
    ^{k}.
  \]
  This design favors skills that are not merely effective for the model from
  which they are distilled, but can also provide reusable procedural guidance to
  other models. The final score of a candidate skill node combines its
  collective quality and transferability:
  \[
    \mathrm{Score}(s_{m,n}) = Q_{m,n}+ \mathrm{Tran}_{m,n}.
  \]
  For each subtask $t_{m}$, CSTS selects the highest-scoring skill node as
  \[
    s_{m}^{\star}= \operatorname*{arg\,max}_{s_{m,n}\in\mathcal{S}_m}\mathrm{Score}
    (s_{m,n}).
  \]
  The selected nodes are then organized according to the task decomposition
  $T=(t_{1},\ldots,t_{M})$. Specifically, the skill path for the complex task
  $T$ is defined as
  \[
    S_{T}^{\star}= (s_{1}^{\star},s_{2}^{\star},\ldots,s_{M}^{\star}),
  \]
  where each $s_{m}^{\star}$ provides procedural guidance for subtask $t_{m}$. In
  this way, $S_{T}^{\star}$ is not a single skill, but an ordered composition of
  subtask-level skills for solving the full task $T$.

  CSTS uses the selected skill path $S_{T}^{\star}$ to augment agent trajectories
  with structured skill guidance, yielding skill-augmented training data. These data
  are subsequently used for supervised fine-tuning (SFT), enabling the policy to
  learn the basic procedural structure before reinforcement learning. For each task
  $T$, we construct an SFT instance
  \[
    (T, S_{T}^{\star}, \tau_{T}^{\star}) \in \mathcal{D}_{\mathrm{SFT}},
  \]
  where $S_{T}^{\star}$ is the selected compositional skill path and
  $\tau_{T}^{\star}$ is the demonstration trajectory assembled from the model
  rollouts that produced the selected skills. The SFT objective is
  \[
    \mathcal{L}_{\mathrm{SFT}}(\theta) = - \mathbb{E}_{(T,S_T^{\star},\tau_T^{\star})\sim\mathcal{D}_{\mathrm{SFT}}}
    \log \pi_{\theta}\left( \tau_{T}^{\star}\mid T,S_{T}^{\star}\right).
  \]

  \subsection{Collective Skill Reinforcement Learning}
  \label{sec:csrl}

  Although CSTS provides structured skill paths for supervised initialization,
  the resulting policy is not explicitly optimized to distinguish which skill-conditioned
  strategies are more effective when multiple candidate skills are available for
  the same subtask. To address this, we introduce \textsc{Collective Skill
  Reinforcement Learning} (CSRL), which extends group-relative policy
  optimization to skill-conditioned rollout groups.

  For each subtask $t_{m}$, given its candidate skill set $\mathcal{S}_{m}$, the
  old policy samples $G$ rollouts conditioned on each skill:
  \[
    \tau_{m,n}^{g}\sim \pi_{\theta_{\mathrm{old}}}(\cdot\mid t_{m},s_{m,n}), \qquad
    g=1,\ldots,G.
  \]
  All rollouts for the same subtask form a collective skill-conditioned group
  \[
    \mathcal{B}_{m}= \{\tau_{m,n}^{g}\mid s_{m,n}\in\mathcal{S}_{m},\; g=1,\ldots
    ,G\}.
  \]
  Each rollout is evaluated by a verifier or reward model as $r_{m,n}^{g}=R(t_{m}
  ,s_{m,n},\tau_{m,n}^{g})$, where the reward may reflect final task success, tool-use
  correctness, intermediate verification, and recovery from execution errors.

  Instead of normalizing rewards only within rollouts generated from the same skill,
  CSRL computes relative advantages over the whole collective group
  $\mathcal{B}_{m}$. Specifically, with group mean $\mu_{m}$ and standard
  deviation $\sigma_{m}$, the advantage is
  \[
    A_{m,n}^{g}= \frac{r_{m,n}^{g}-\mu_{m}}{\sigma_{m}+\delta}, \qquad \mu_{m}=\frac{1}{|\mathcal{B}_{m}|}
    \sum_{\tau_{m,n}^{g}\in\mathcal{B}_m}r_{m,n}^{g}, \quad \sigma_{m}=\operatorname{Std}
    \{r_{m,n}^{g}\mid \tau_{m,n}^{g}\in\mathcal{B}_{m}\}.
  \]
  This cross-skill normalization makes each rollout compete with trajectories generated
  under alternative skills for the same subtask, encouraging the policy to favor
  more effective skill-conditioned strategies.

  We then optimize the policy with a GRPO-style clipped objective. For each rollout
  $\tau_{m,n}^{g}$, let $L_{m,n}^{g}$ denote its length, and let
  \[
    h_{m,n,<\ell}^{g}= \{(\psi_{m,n,j}^{g},a_{m,n,j}^{g},o_{m,n,j}^{g})\}_{j<\ell}
  \]
  denote the interaction history before step $\ell$, including previous
  reasoning states, actions, and observations. The action-level probability ratio
  is then defined as
  \[
    \rho_{m,n,\ell}^{g}(\theta) = \frac{ \pi_{\theta}\left( a_{m,n,\ell}^{g}\mid
    t_{m},s_{m,n},h_{m,n,<\ell}^{g}\right) }{ \pi_{\theta_{\mathrm{old}}}\left(
    a_{m,n,\ell}^{g}\mid t_{m},s_{m,n},h_{m,n,<\ell}^{g}\right) }.
  \]
  For compactness, we define the clipped surrogate term as
  \[
    u_{m,n,\ell}^{g}(\theta) = \min \left( \rho_{m,n,\ell}^{g}(\theta)A_{m,n}^{g}
    , \operatorname{clip}(\rho_{m,n,\ell}^{g}(\theta),1-\epsilon,1+\epsilon)A_{m,n}
    ^{g}\right).
  \]
 
  Then, the CSRL loss is defined as follows to train our final model, OpenClaw-Skill:
  \[
    \mathcal{L}_{\mathrm{CSRL}}(\theta) = - \mathbb{E}_{t_m,\mathcal{B}_m}\left[
    \frac{1}{|\mathcal{B}_{m}|}\sum_{\tau_{m,n}^{g}\in\mathcal{B}_m}\frac{1}{L_{m,n}^{g}}
    \sum_{\ell=1}^{L_{m,n}^{g}}u_{m,n,\ell}^{g}(\theta) \right].
  \]

  Through this objective, CSRL increases the likelihood of actions from high-reward
  rollouts whose advantages are positive relative to the collective skill-conditioned
  group, while suppressing ineffective skill-conditioned behaviors. In this way,
  CSTS constructs the collective skill space, and CSRL converts the relative effectiveness
  of different skills into a direct policy optimization signal.

  \section{Experiments}
  \label{sec:experiments}

  In this section, we first provide implementation details, and then present
  main results on two real-world agentic benchmarks, i.e., \textsc{QwenClawBench}
  and \textsc{PinchBench}, which cover a broad range of long-horizon agentic
  tasks involving tool use, file operations, code execution, web interaction,
  multi-step decision making, etc. In final, we provide ablation studies and discussion
  on our methods.

  \subsection{Implementation Details}
  \label{sec:implementation}

  We conduct experiments based on four popular backbones, including Qwen3-4B,
  Qwen3-8B, Qwen3.5-4B, and Qwen3.5-9B.
  During training, CSTS first decomposes each task into subtasks, collects multi-agent
  rollouts, synthesizes candidate skill nodes, and constructs skill-augmented trajectories.
  Using CSTS, we collect 2K high-quality SFT examples and fine-tune each model for
  2 epochs on 8 H100 GPUs with a learning rate of $5\times10^{-6}$.
  We evaluate on \textsc{QwenClawBench} and \textsc{PinchBench}, reporting
  category-level and overall scores for the former, and best-run and mean success
  rates for the latter.

  \subsection{Main Results}
  \label{sec:main_results}

  \begin{table*}
    [t]
    \caption{ Comparis In this section, we first provide implementation detail, and
    then present main results, trained with CSTS-generated data and CSRL, against
    other models on \textsc{QwenClawBench} across multiple task categories~\citep{qwenclawbench,qwen36_35b_a3b}.
    }
    \label{tab:qwenclawbench}
    \centering
    \small
    \setlength{\tabcolsep}{3.0pt}
    \begin{tabular}{lccccccccl@{}}
      \toprule Model                                                                               & WAO           & SOA           & KMM           & FQT           & DAM           & SVM           & CS            & RIR           & Overall                                                          \\
      \midrule \multicolumn{10}{l}{\textit{Closed-source models}}                                   \\
      \midrule                                                                   
      Claude Opus 4.6~\citep{anthropic2026claudeopus46}                                                                              & 57.8          & 57.3          & 55.6          & 48.5          & 54.3          & 76.8          & 67.3          & 71.0          & 59.5                                                             \\
      Qwen3.6-Plus~\citep{qwen2026qwen36plus}                                                                                 & 56.2          & 52.2          & 55.1          & 46.6          & 56.1          & 76.9          & 62.6          & 67.5          & 57.4                                                             \\
      GPT-5.4~\citep{openai2026gpt54}                                                                                      & 55.2          & 54.3          & 57.9          & 38.4          & 52.6          & 70.6          & 71.1          & 63.7          & 56.7                                                             \\
      MiMo-V2-Pro~\citep{mimo2026v2pro} & 57.1          & 49.1          & 57.3          & 46.4          & 53.4          & 74.0          & 60.6          & 64.9          & 56.5                                                             \\
      \addlinespace[0.25em] \midrule \multicolumn{10}{l}{\textit{Open-source / open-weight models}} \\
      \midrule \addlinespace[0.15em] GLM-5.1~\citep{zai2026glm51}                                                       & 63.1          & 53.3          & 58.8          & 39.4          & 49.4          & 75.7          & 66.9          & 70.8          & 58.7                                                             \\
      Kimi-K2.5~\citep{moonshot2026kimi25}                                                                                    & 51.6          & 51.5          & 50.7          & 39.3          & 46.3          & 62.2          & 60.2          & 59.8          & 51.9                                                             \\
      DeepSeek-V3.2-Thinking~\citep{deepseek2025v32}                                                                       & 45.5          & 46.8          & 49.7          & 35.6          & 57.1          & 63.8          & 63.7          & 60.7          & 50.7                                                             \\
      MiniMax-M2.7~\citep{minimax2026m27}                                                                                 & 46.6          & 46.5          & 53.7          & 40.6          & 54.2          & 62.7          & 51.7          & 58.7          & 50.5                                                             \\
      Gemma4-31B~\citep{google2026gemma4}                                                                                   & --            & --            & --            & --            & --            & --            & --            & --            & 41.7                                                             \\
      Gemma4-26B-A4B~\cite{google2026gemma4}                                                                               & --            & --            & --            & --            & --            & --            & --            & --            & 38.7                                                             \\
      \cmidrule(lr){1-10}                                                              
      Qwen3-4B~\citep{qwen2025qwen3}                                                                                     & 6.6           & 10.8          & 7.1           & 0.8           & 7.2           & 2.5           & 6.3           & 13.2          & 7.0                                                              \\
      \rowcolor{mygray} OpenClaw-Skill-Qwen3 4B                                                    & 13.5          & 10.2          & 21.9          & 2.9           & 10.1          & 11.3          & 23.3          & 7.2           & 12.8$^{\color{teal}{\textbf{\scriptsize5.8}\uparrow}}$           \\
      Qwen3-8B~\citep{qwen2025qwen3}                                                                                     & 17.3          & 7.9           & 13.0          & 2.0           & 14.2          & 13.1          & 9.3           & 10.7          & 11.5                                                             \\
      \rowcolor{mygray} OpenClaw-Skill-Qwen3 8B                                                    & 12.7          & 15.9          & 18.2          & 5.8           & 9.4           & 22.0          & 34.3          & 13.3          & 15.8$^{\color{teal}{\textbf{\scriptsize4.3}\uparrow}}$           \\
      Qwen3.5-4B~\citep{qwen2026qwen35}                                                                                & 24.3          & 28.8          & 34.2          & 15.4          & 34.3          & 48.3          & 55.4          & 24.4          & 31.5                                                             \\
      \rowcolor{mygray} OpenClaw-Skill 4B                                                          & \textbf{35.8} & \textbf{46.1} & \textbf{46.7} & \textbf{7.8}  & \textbf{33.2} & \textbf{51.5} & \textbf{61.0} & \textbf{54.1} & \textbf{41.2}$^{\color{teal}{\textbf{\scriptsize9.7}\uparrow}}$  \\
      Qwen3.5-9B~\citep{qwen2026qwen35}                                                                                     & 26.6          & 39.9          & 58.8          & 15.4          & 19.9          & 33.2          & 30.2          & 44.8          & 34.5                                                             \\
      \rowcolor{mygray} OpenClaw-Skill 9B                                                          & \textbf{32.3} & \textbf{43.3} & \textbf{60.2} & \textbf{14.2} & \textbf{28.4} & \textbf{70.9} & \textbf{78.4} & \textbf{50.1} & \textbf{44.9}$^{\color{teal}{\textbf{\scriptsize10.4}\uparrow}}$ \\
      \bottomrule
    \end{tabular}
  \end{table*}

  \paragraph{Results on QwenClawBench.}
  Table~\ref{tab:qwenclawbench} shows that \textsc{OpenClaw-Skill} brings consistent
  overall gains across all evaluated Qwen backbones. The overall score improves
  by 5.8 points for Qwen3-4B, 4.3 points for Qwen3-8B, 9.7 points for Qwen3.5-4B,
  and 10.4 points for Qwen3.5-9B. Notably, OpenClaw-Skill 4B and OpenClaw-Skill 9B,
  which are built on Qwen3.5 backbones by default, reach 41.2 and 44.9 overall, respectively.

  The gains are most evident on categories involving long-horizon tool use and execution
  feedback. For instance, OpenClaw-Skill 9B improves SVM from 33.2 to 70.9 and CS
  from 30.2 to 78.4, while OpenClaw-Skill 4B improves RIR from 24.4 to 54.1. This
  suggests that CSTS-generated skills and CSRL improve the model's ability to follow
  procedural guidance, verify intermediate states, and recover from execution errors.

  \begin{table*}
    [t]
    \caption{
    Comparison of \textsc{OpenClaw-Skill}, trained with CSTS-generated data and
    CSRL, against other models on \textsc{PinchBench}~\citep{pinchbench} over 23-task
    original version and its 123-task expanded version. }
    \label{tab:pinchbench}
    \vspace{0.3em}
    \centering
    \small
    \setlength{\tabcolsep}{6pt}
    \renewcommand{\arraystretch}{1.08}
    \begin{tabular}{@{}lcccc@{}}
      \toprule Model                                                        & \multicolumn{2}{c}{\textsc{PinchBench} (23 tasks)} & \multicolumn{2}{c}{\textsc{PinchBench} (123 tasks)} \\
      \cmidrule(lr){2-3} \cmidrule(lr){4-5}                                 & Best (\%)                                          & Average (\%)                                       & Best (\%)     & Average (\%)  \\
      \midrule \multicolumn{5}{l}{\textit{Closed-source models}}             \\
      \midrule Claude-Opus-4.6~\citep{anthropic2026claudeopus46}                                              & 93.3                                               & 81.6                                               & 88.9          & 67.5          \\
      Claude-Haiku-4.5~\citep{anthropic2025claudehaiku45} & 89.5 & 77.4 & 90.3 & 58.9 \\
      GPT-5.4~\citep{openai2026gpt54}                                                               & 90.5                                               & 79.4                                               & --            & --            \\
      Gemini-3.1-Pro-Preview~\citep{google2026gemini31pro}                                                & 86.7                                               & 77.5                                               & 81.7          & 80.2          \\
      MiMo-V2-Pro~\citep{mimo2026v2pro} & 87.4 & 80.4 & -- & -- \\
      \midrule \multicolumn{5}{l}{\textit{Open-source / open-weight models}} \\
      \midrule MiniMax-M2.7~\citep{minimax2026m27}                                                 & 89.8                                               & 82.8                                               & --            & --            \\
      MiMo-V2-Flash~\citep{mimo2025v2flash}                                                         & 88.8                                               & 69.7                                               & --            & --            \\
      Qwen3.6-Plus~\citep{qwen2026qwen36plus}                                                          & 63.9                                               & 63.3                                               & 82.8          & 62.6          \\
      GPT-OSS-120B~\citep{openai2025gptoss} & 67.1 & 52.0 & 47.4 & 44.5 \\
      GPT-OSS-20B~\citep{openai2025gptoss}                                                           & 66.0                                               & 50.3                                               & 41.8          & 37.5          \\
      Nemotron-3-Super-120B-A12B~\citep{nvidia2025nemotron3}                                            & --                                                 & --                                                 & 50.6          & 38.6          \\
      Llama 4 Maverick~\citep{meta2025llama4}                                                      & 46.1                                               & 34.8                                               & --            & --            \\
      \midrule                                                             
      Qwen3-4B~\citep{qwen2025qwen3}                                                              & 45.2                                               & 31.8                                               & 22.4          & 13.6          \\
      \rowcolor{mygray} OpenClaw-Skill-Qwen3-4B                             & 64.5                                               & 47.9                                               & 31.1          & 20.8          \\
      Qwen3-8B~\citep{qwen2025qwen3}                                                              & 49.8                                               & 35.4                                               & 27.9          & 18.3          \\
      \rowcolor{mygray} OpenClaw-Skill-Qwen3 8B                             & 64.9                                               & 49.2                                               & 33.0          & 22.5          \\
      Qwen3.5-4B~\citep{qwen2026qwen35}                                                            & 71.0                                               & 55.7                                               & 60.9          & 45.9          \\
      \rowcolor{mygray} OpenClaw-Skill 4B                                   & \textbf{71.5}                                      & \textbf{56.4}                                      & \textbf{61.4} & \textbf{47.6} \\
      Qwen3.5-9B~\citep{qwen2026qwen35}                                                            & 67.5                                               & 53.8                                               & 61.1          & 47.1          \\
      \rowcolor{mygray} OpenClaw-Skill 9B                                   & \textbf{72.8}                                      & \textbf{58.9}                                      & \textbf{68.2} & \textbf{53.6} \\
      \bottomrule
    \end{tabular}
  \end{table*}

  \paragraph{Results on PinchBench.}
  Table~\ref{tab:pinchbench} reports results on \textsc{PinchBench}. \textsc{OpenClaw-Skill}
  consistently improves the corresponding Qwen backbones on both the 23-task
  early benchmark and the 123-task expanded benchmark. On the 23-task setting,
  OpenClaw-Skill 9B improves Qwen3.5-9B from 67.5 to 72.8 in best success rate and
  from 53.8 to 58.9 in average success rate. On the 123-task setting, the gain is
  more evident: OpenClaw-Skill 9B improves the best score from 61.1 to 68.2 and
  the average score from 47.1 to 53.6. The 4B models also show consistent gains.
  OpenClaw-Skill 4B improves Qwen3.5-4B from 60.9 to 61.4 in best score and from
  45.9 to 47.6 in average score on the 123-task setting. For the smaller Qwen3
  backbones, OpenClaw-Skill also improves the average score from 13.6 to 20.8 for
  Qwen3-4B and from 18.3 to 22.5 for Qwen3-8B. These results suggest that CSTS-generated
  skills and CSRL improve both peak performance and average execution robustness
  across different PinchBench versions.

  \subsection{Ablation Study}
  \label{sec:ablation} Table~\ref{tab:ablation} studies the contribution of each
  component in \textsc{OpenClaw-Skill} using Qwen3.5-9B as the backbone. Starting
  from the base model, adding CSN-Gen improves the overall score from 34.5 to
  39.8, showing that diverse skills distilled from collective trajectories provide
  useful procedural supervision. Further incorporating CSN-Assess increases the
  score to 42.8, indicating that multi-judge quality assessment and cross-model transferability
  evaluation help filter noisy or less reusable skill nodes. Finally, adding CSRL
  further improves the score to 44.9, demonstrating that reinforcement learning
  over skill-conditioned rollout groups provides additional policy improvement
  beyond skill-based SFT. Overall, the ablation confirms that both CSTS-based
  skill construction and CSRL-based policy optimization contribute to the final
  performance gain.

  \begin{table}[t]
    \centering
    \small
    \caption{ Ablation study of \textsc{OpenClaw-Skill} on \textsc{QwenClawBench}.
    }
    \label{tab:ablation}
    \setlength{\tabcolsep}{6pt}
    \renewcommand{\arraystretch}{1.08}
    \begin{tabular}{lcccc}
      \toprule \multirow{2}{*}{Setting} & \multicolumn{2}{c}{CSTS} & \multirow{2}{*}{CSRL} & \multirow{2}{*}{Overall} \\
      \cmidrule(lr){2-3}                & CSN-Gen                  & CSN-Assess            &                         &               \\
      \midrule Qwen3.5-9B               & --                       & --                    & --                      & 34.5          \\
      + CSN-Gen                         & \checkmark               & --                    & --                      & 39.8          \\
      + CSN-Gen + CSN-Assess            & \checkmark               & \checkmark            & --                      & 42.8          \\
      \textsc{OpenClaw-Skill}           & \checkmark               & \checkmark            & \checkmark              & \textbf{44.9} \\
      \bottomrule
    \end{tabular}
  \end{table}

  \section{Conclusion}
  In this paper, we present \textsc{OpenClaw-Skill}, a framework with automatic skill construction and skill-augmented training to enhance LLM agents on complex tasks on real-world sytems like OpenClaw. 
  At the core of our framework is Collective Skill Tree Search (CSTS), a novel tree-search-based skill construction framework that builds structured, diverse, and generalizable skills for LLM agents, improving their ability to solve sophisticated real-world tasks. CSTS operates through two iterative phases: \textit{Collective Skill Node Generation} and \textit{Collective Skill Node Assessment}. Together, these phases leverage collective intelligence to iteratively search, identify, and compose effective skills into cohesive and reliable \textit{trees of skills}.
  Furthermore, we introduce \textit{Collective Skill Reinforcement Learning}, which further optimizes the policy over skill-conditioned rollout groups and encourages the model to favor more effective procedural strategies. Extensive experiments on \textsc{QwenClawBench} and \textsc{PinchBench}, together with ablation studies and qualitative analyses, demonstrate that \textsc{OpenClaw-Skill} consistently achieves strong performance on diverse long-horizon tasks involving tool use, file operations, web interaction, and execution feedback.

  { \small \bibliographystyle{unsrt} \bibliography{references} }

@misc{anthropic2025agentSkills,
  title        = {Equipping agents for the real world with Agent Skills},
  author       = {{Anthropic}},
  year         = {2025},
  month        = oct,
  howpublished = {\url{https://www.anthropic.com/engineering/equipping-agents-for-the-real-world-with-agent-skills}},
  note         = {Anthropic Engineering Blog}
}

@article{xia2026skillrl,
  title={Skillrl: Evolving agents via recursive skill-augmented reinforcement learning},
  author={Xia, Peng and Chen, Jianwen and Wang, Hanyang and Liu, Jiaqi and Zeng, Kaide and Wang, Yu and Han, Siwei and Zhou, Yiyang and Zhao, Xujiang and Chen, Haifeng and others},
  journal={arXiv preprint arXiv:2602.08234},
  year={2026}
}

@misc{qwenclawbench,
    title = {{QwenClawBench}: Real-user-distribution benchmark for OpenClaw agents},
    url = {github.com/SKYLENAGE-AI/QwenClawBench},
    author = {{Qwen Team} and {Alibaba Data}},
    month = {April},
    year = {2026}
}

@misc{pinchbench,
  title        = {{PinchBench: OpenClaw LLM Model Benchmarking}},
  author       = {{Kilo Code}},
  year         = {2026},
  howpublished = {\url{https://pinchbench.com/}},
  note         = {Accessed: 2026-05-07}
}

@misc{qwen36_35b_a3b,
  title        = {{Qwen3.6-35B-A3B}},
  author       = {{Qwen Team}},
  year         = {2026},
  howpublished = {\url{https://qwen.ai/blog?id=qwen3.6-35b-a3b}},
  note         = {Accessed: 2026-05-07}
}

@inproceedings{react,
  title = {{ReAct}: Synergizing Reasoning and Acting in Language Models},
  author = {Yao, Shunyu and Zhao, Jeffrey and Yu, Dian and Du, Nan and Shafran, Izhak and Narasimhan, Karthik R. and Cao, Yuan},
  booktitle = {The Eleventh International Conference on Learning Representations},
  year = {2023},
  url = {https://openreview.net/forum?id=WE_vluYUL-X}
}

@inproceedings{toolformer,
  title = {Toolformer: Language Models Can Teach Themselves to Use Tools},
  author = {Schick, Timo and Dwivedi-Yu, Jane and Dessi, Roberto and Raileanu, Roberta and Lomeli, Maria and Hambro, Eric and Zettlemoyer, Luke and Cancedda, Nicola and Scialom, Thomas},
  booktitle = {Advances in Neural Information Processing Systems},
  volume = {36},
  pages = {68539--68551},
  year = {2023},
  url = {https://openreview.net/forum?id=Yacmpz84TH}
}

@inproceedings{webvoyager,
  title = {WebVoyager: Building an End-to-End Web Agent with Large Multimodal Models},
  author = {He, Hongliang and Yao, Wenlin and Ma, Kaixin and Yu, Wenhao and Dai, Yong and Zhang, Hongming and Lan, Zhenzhong and Yu, Dong},
  booktitle = {Proceedings of the 62nd Annual Meeting of the Association for Computational Linguistics (Volume 1: Long Papers)},
  pages = {6864--6890},
  address = {Bangkok, Thailand},
  publisher = {Association for Computational Linguistics},
  year = {2024},
  doi = {10.18653/v1/2024.acl-long.371},
  url = {https://aclanthology.org/2024.acl-long.371/}
}

@inproceedings{osworld,
  title = {OSWorld: Benchmarking Multimodal Agents for Open-Ended Tasks in Real Computer Environments},
  author = {Xie, Tianbao and Zhang, Danyang and Chen, Jixuan and Li, Xiaochuan and Zhao, Siheng and Cao, Ruisheng and Hua, Toh Jing and Cheng, Zhoujun and Shin, Dongchan and Lei, Fangyu and Liu, Yitao and Xu, Yiheng and Zhou, Shuyan and Savarese, Silvio and Xiong, Caiming and Zhong, Victor and Yu, Tao},
  booktitle = {Advances in Neural Information Processing Systems},
  volume = {37},
  pages = {52040--52094},
  note = {Datasets and Benchmarks Track},
  year = {2024},
  doi = {10.52202/079017-1650},
  url = {https://proceedings.neurips.cc/paper_files/paper/2024/hash/5d413e48f84dc61244b6be550f1cd8f5-Abstract-Datasets_and_Benchmarks_Track.html}
}

@misc{openclaw,
  title = {OpenClaw},
  author = {{OpenClaw}},
  year = {2026},
  url = {https://docs.openclaw.ai/},
  note = {Official documentation and repository, accessed 2026-05-07}
}

@misc{deerflow,
  title = {DeerFlow: Deep Exploration and Efficient Research Flow},
  author = {{ByteDance}},
  year = {2026},
  url = {https://github.com/bytedance/deer-flow},
  note = {Official repository, accessed 2026-05-07}
}

@misc{hermes,
  title = {Hermes Agent},
  author = {{Nous Research}},
  year = {2026},
  url = {https://github.com/NousResearch/hermes-agent},
  note = {Official repository, accessed 2026-05-07}
}

@article{clawgym,
  title = {ClawGym: A Scalable Framework for Building Effective Claw Agents},
  author = {Bai, Fei and Song, Huatong and Sun, Shuang and Cheng, Daixuan and Yang, Yike and Hao, Chuan and Li, Renyuan and Chang, Feng and Wei, Yuan and Tao, Ran and Dai, Bryan and Yang, Jian and Zhao, Wayne Xin},
  journal = {arXiv preprint arXiv:2604.26904},
  year = {2026},
  url = {https://arxiv.org/abs/2604.26904}
}

@article{openclawrl,
  title = {OpenClaw-RL: Train Any Agent Simply by Talking},
  author = {Wang, Yinjie and Chen, Xuyang and Jin, Xiaolong and Wang, Mengdi and Yang, Ling},
  journal = {arXiv preprint arXiv:2603.10165},
  year = {2026},
  url = {https://arxiv.org/abs/2603.10165}
}

@article{skillsbench,
  title = {SkillsBench: Benchmarking How Well Agent Skills Work Across Diverse Tasks},
  author = {Li, Xiangyi and Chen, Wenbo and Liu, Yimin and Zheng, Shenghan and Chen, Xiaokun and He, Yifeng and Li, Yubo and You, Bingran and Shen, Haotian and Sun, Jiankai and Wang, Shuyi and Li, Binxu and Zeng, Qunhong and Wang, Di and Zhao, Xuandong and Wang, Yuanli and Ben Chaim, Roey and Di, Zonglin and Gao, Yipeng and He, Junwei and He, Yizhuo and Jing, Liqiang and Kong, Luyang and Lan, Xin and Li, Jiachen and Li, Songlin and Li, Yijiang and Lin, Yueqian and Liu, Xinyi and Liu, Xuanqing and Lyu, Haoran and Ma, Ze and Wang, Bowei and Wang, Runhui and Wang, Tianyu and Ye, Wengao and Zhang, Yue and Xing, Hanwen and Xue, Yiqi and Dillmann, Steven and Lee, Han-chung},
  journal = {arXiv preprint arXiv:2602.12670},
  year = {2026},
  url = {https://arxiv.org/abs/2602.12670}
}

@article{skillcraft,
  title = {SkillCraft: Can LLM Agents Learn to Use Tools Skillfully?},
  author = {Chen, Shiqi and Gai, Jingze and Zhou, Ruochen and Zhang, Jinghan and Zhu, Tongyao and Li, Junlong and Wang, Kangrui and Wang, Zihan and Chen, Zhengyu and Kaleb, Klara and Miao, Ning and Gao, Siyang and Lu, Cong and Li, Manling and He, Junxian and Teh, Yee Whye},
  journal = {arXiv preprint arXiv:2603.00718},
  year = {2026},
  url = {https://arxiv.org/abs/2603.00718}
}

@article{skillrl,
  title = {SkillRL: Evolving Agents via Recursive Skill-Augmented Reinforcement Learning},
  author = {Xia, Peng and Chen, Jianwen and Wang, Hanyang and Liu, Jiaqi and Zeng, Kaide and Wang, Yu and Han, Siwei and Zhou, Yiyang and Zhao, Xujiang and Chen, Haifeng and Zheng, Zeyu and Xie, Cihang and Yao, Huaxiu},
  journal = {arXiv preprint arXiv:2602.08234},
  year = {2026},
  url = {https://arxiv.org/abs/2602.08234}
}

@article{evoskill,
  title = {EvoSkill: Automated Skill Discovery for Multi-Agent Systems},
  author = {Alzubi, Salaheddin and Provenzano, Noah and Bingham, Jaydon and Chen, Weiyuan and Vu, Tu},
  journal = {arXiv preprint arXiv:2603.02766},
  year = {2026},
  url = {https://arxiv.org/abs/2603.02766}
}

@article{trace2skill,
  title = {Trace2Skill: Distill Trajectory-Local Lessons into Transferable Agent Skills},
  author = {Ni, Jingwei and Liu, Yihao and Liu, Xinpeng and Sun, Yutao and Zhou, Mengyu and Cheng, Pengyu and Wang, Dexin and Zhao, Erchao and Jiang, Xiaoxi and Jiang, Guanjun},
  journal = {arXiv preprint arXiv:2603.25158},
  year = {2026},
  url = {https://arxiv.org/abs/2603.25158}
}

@article{coevoskills,
  title = {CoEvoSkills: Self-Evolving Agent Skills via Co-Evolutionary Verification},
  author = {Zhang, Hanrong and Fan, Shicheng and Zou, Henry Peng and Chen, Yankai and Wang, Zhenting and Zhou, Jiayu and Li, Chengze and Huang, Wei-Chieh and Yao, Yifei and Zheng, Kening and Liu, Xue and Li, Xiaoxiao and Yu, Philip S.},
  journal = {arXiv preprint arXiv:2604.01687},
  year = {2026},
  url = {https://arxiv.org/abs/2604.01687}
}

@article{gpt5,
  title   = {OpenAI GPT-5 System Card},
  author  = {{OpenAI}},
  journal = {arXiv preprint arXiv:2601.03267},
  year    = {2026},
  url     = {https://arxiv.org/abs/2601.03267}
}

@misc{openai2026gpt54,
  title        = {Introducing {GPT-5.4}},
  author       = {{OpenAI}},
  year         = {2026},
  howpublished = {\url{https://openai.com/index/introducing-gpt-5-4/}},
  note         = {Accessed: 2026-05-07}
}

@misc{zai2026glm51,
  title        = {{GLM-5.1}: Towards Long-Horizon Tasks},
  author       = {{Z.AI}},
  year         = {2026},
  howpublished = {\url{https://z.ai/blog/glm-5.1}},
  note         = {Accessed: 2026-05-07}
}

@misc{moonshot2026kimi25,
  title        = {{Kimi K2.5}: Visual Agentic Intelligence},
  author       = {{Moonshot AI}},
  year         = {2026},
  howpublished = {\url{https://www.kimi.com/blog/kimi-k2-5}},
  note         = {Accessed: 2026-05-07}
}

@misc{deepseek2025v32,
  title        = {{DeepSeek-V3.2} Release},
  author       = {{DeepSeek-AI}},
  year         = {2025},
  howpublished = {\url{https://api-docs.deepseek.com/news/news251201}},
  note         = {Accessed: 2026-05-07}
}

@misc{minimax2026m27,
  title        = {{MiniMax M2.7}: Early Echoes of Self-Evolution},
  author       = {{MiniMax}},
  year         = {2026},
  howpublished = {\url{https://www.minimax.io/news/minimax-m27-en}},
  note         = {Accessed: 2026-05-07}
}

@misc{google2026gemma4,
  title        = {{Gemma 4}: Byte for Byte, the Most Capable Open Models},
  author       = {{Google}},
  year         = {2026},
  howpublished = {\url{https://blog.google/innovation-and-ai/technology/developers-tools/gemma-4/}},
  note         = {Accessed: 2026-05-07}
}

@misc{mimo2026v2pro,
  title        = {{MiMo-V2-Pro}},
  author       = {{Xiaomi MiMo Team}},
  year         = {2026},
  howpublished = {\url{https://mimo.xiaomi.com/mimo-v2-pro}},
  note         = {Accessed: 2026-05-07}
}

@misc{qwen2026qwen35,
  title        = {{Qwen3.5}: Towards Native Multimodal Agents},
  author       = {{Qwen Team}},
  year         = {2026},
  howpublished = {\url{https://qwen.ai/blog?id=qwen3.5}},
  note         = {Accessed: 2026-05-07}
}

@misc{anthropic2025claudehaiku45,
  title        = {Introducing {Claude Haiku 4.5}},
  author       = {{Anthropic}},
  year         = {2025},
  howpublished = {\url{https://www.anthropic.com/news/claude-haiku-4-5}},
  note         = {Accessed: 2026-05-07}
}

@misc{google2026gemini31pro,
  title        = {{Gemini 3.1 Pro}: A Smarter Model for Your Most Complex Tasks},
  author       = {{Google}},
  year         = {2026},
  howpublished = {\url{https://blog.google/innovation-and-ai/models-and-research/gemini-models/gemini-3-1-pro/}},
  note         = {Accessed: 2026-05-07}
}

@misc{mimo2025v2flash,
  title        = {{MiMo-V2-Flash}},
  author       = {{Xiaomi MiMo Team}},
  year         = {2025},
  howpublished = {\url{https://github.com/XiaomiMiMo/MiMo-V2-Flash}},
  note         = {Accessed: 2026-05-07}
}

@article{openai2025gptoss,
  title        = {{gpt-oss-120b \& gpt-oss-20b} Model Card},
  author       = {Agarwal, S. and others},
  journal      = {arXiv preprint arXiv:2508.10925},
  year         = {2025}
}

@article{nvidia2025nemotron3,
  title        = {{NVIDIA Nemotron 3}: Efficient and Open Intelligence},
  author       = {Blakeman, A. and others},
  journal      = {arXiv preprint arXiv:2512.20856},
  year         = {2025}
}

@misc{meta2025llama4,
  title        = {The {Llama 4} Herd: The Beginning of a New Era of Natively Multimodal AI Innovation},
  author       = {{Meta AI}},
  year         = {2025},
  howpublished = {\url{https://ai.meta.com/blog/llama-4-multimodal-intelligence/}},
  note         = {Accessed: 2026-05-07}
}

@inproceedings{treeofthought,
  author    = {Shunyu Yao and Dian Yu and Jeffrey Zhao and Izhak Shafran and Tom Griffiths and Yuan Cao and Karthik Narasimhan},
  editor    = {Alice Oh and Tristan Naumann and Amir Globerson and Kate Saenko and Moritz Hardt and Sergey Levine},
  title     = {Tree of Thoughts: Deliberate Problem Solving with Large Language Models},
  booktitle = {Advances in Neural Information Processing Systems 36: Annual Conference on Neural Information Processing Systems 2023, NeurIPS 2023, New Orleans, LA, USA, December 10--16, 2023},
  volume    = {36},
  year      = {2023},
  publisher = {Neural Information Processing Systems Foundation},
  url       = {https://papers.nips.cc/paper_files/paper/2023/hash/271db9922b8d1f4dd7aaef84ed5ac703-Abstract-Conference.html}
}

@inproceedings{lats,
  title     = {Language Agent Tree Search Unifies Reasoning, Acting, and Planning in Language Models},
  author    = {Zhou, Andy and Yan, Kai and Shlapentokh-Rothman, Michal and Wang, Haohan and Wang, Yu-Xiong},
  booktitle = {Proceedings of the 41st International Conference on Machine Learning},
  pages     = {62138--62160},
  year      = {2024},
  editor    = {Salakhutdinov, Ruslan and Kolter, Zico and Heller, Katherine and Weller, Adrian and Oliver, Nuria and Scarlett, Jonathan and Berkenkamp, Felix},
  volume    = {235},
  series    = {Proceedings of Machine Learning Research},
  month     = {21--27 Jul},
  publisher = {PMLR},
  url       = {https://proceedings.mlr.press/v235/zhou24r.html}
}

@article{mulberry,
  author      = {Huanjin Yao and Jiaxing Huang and Wenhao Wu and Jingyi Zhang and Yibo Wang and Shunyu Liu and Yingjie Wang and Yuxin Song and Haocheng Feng and Li Shen and Dacheng Tao},
  title       = {Mulberry: Empowering {MLLM} with o1-like Reasoning and Reflection via Collective Monte Carlo Tree Search},
  journal     = {CoRR},
  volume      = {abs/2412.18319},
  year        = {2024},
  url         = {https://doi.org/10.48550/arXiv.2412.18319},
  doi         = {10.48550/ARXIV.2412.18319},
  eprinttype  = {arXiv},
  eprint      = {2412.18319}
}

@article{treesearchagents,
  author  = {Jing Yu Koh and Stephen Marcus McAleer and Daniel Fried and Ruslan Salakhutdinov},
  title   = {Tree Search for Language Model Agents},
  journal = {Transactions on Machine Learning Research},
  year    = {2025},
  url     = {https://openreview.net/forum?id=QF0N3x2XVm}
}

@article{silver2017mastering,
  title={Mastering the game of go without human knowledge},
  author={Silver, David and Schrittwieser, Julian and Simonyan, Karen and Antonoglou, Ioannis and Huang, Aja and Guez, Arthur and Hubert, Thomas and Baker, Lucas and Lai, Matthew and Bolton, Adrian and others},
  journal={nature},
  volume={550},
  number={7676},
  pages={354--359},
  year={2017},
  publisher={Nature Publishing Group}
}

@article{ye2021mastering,
  title={Mastering atari games with limited data},
  author={Ye, Weirui and Liu, Shaohuai and Kurutach, Thanard and Abbeel, Pieter and Gao, Yang},
  journal={Advances in neural information processing systems},
  volume={34},
  pages={25476--25488},
  year={2021}
}

@misc{claude_sonnet_4,
  author= {Anthropic},
  title={Claude-4-Sonnet},
  year={2025},
}

@misc{anthropic2026claudeopus46,
  title        = {Introducing Claude Opus 4.6},
  author       = {{Anthropic}},
  year         = {2026},
  howpublished = {\url{https://www.anthropic.com/news/claude-opus-4-6}},
  note         = {Accessed: 2026-05-07}
}

@misc{qwen2026qwen36plus,
  title        = {{Qwen3.6-Plus}: Towards Real World Agents},
  author       = {{Qwen Team}},
  year         = {2026},
  howpublished = {\url{https://qwen.ai/blog?id=qwen3.6}},
  note         = {Accessed: 2026-05-07}
}

@article{gemini2.5,
  title={Gemini 2.5: Pushing the frontier with advanced reasoning, multimodality, long context, and next generation agentic capabilities},
  author={Comanici, Gheorghe and Bieber, Eric and Schaekermann, Mike and Pasupat, Ice and Sachdeva, Noveen and Dhillon, Inderjit and Blistein, Marcel and Ram, Ori and Zhang, Dan and Rosen, Evan and others},
  journal={arXiv preprint arXiv:2507.06261},
  year={2025}
}

@article{vicuna,
  title={Vicuna: An open-source chatbot impressing gpt-4 with 90\%* chatgpt quality},
  author={Chiang, Wei-Lin and Li, Zhuohan and Lin, Zi and Sheng, Ying and Wu, Zhanghao and Zhang, Hao and Zheng, Lianmin and Zhuang, Siyuan and Zhuang, Yonghao and Gonzalez, Joseph E and others},
  journal={See https://vicuna. lmsys. org (accessed 14 April 2023)},
  volume={2},
  number={3},
  pages={6},
  year={2023}
}

@article{xskill,
  title={XSkill: Continual Learning from Experience and Skills in Multimodal Agents},
  author={Jiang, Guanyu and Su, Zhaochen and Qu, Xiaoye and Fung, Yi R.},
  journal={arXiv preprint arXiv:2603.12056},
  year={2026}
}

@article{skill0,
  title={SKILL0: In-Context Agentic Reinforcement Learning for Skill Internalization},
  author={Lu, Zhengxi and Yao, Zhiyuan and Wu, Jinyang and Han, Chengcheng and Gu, Qi and Cai, Xunliang and Lu, Weiming and Xiao, Jun and Zhuang, Yueting and Shen, Yongliang},
  journal={arXiv preprint arXiv:2604.02268},
  year={2026}
}

@article{plaat2025agentic,
  title={Agentic Large Language Models, a survey},
  author={Plaat, Aske and van Duijn, Max and van Stein, Niki and Preuss, Mike and van der Putten, Peter and Batenburg, Kees Joost},
  journal={arXiv preprint arXiv:2503.23037},
  year={2025}
}

@article{luo2025llmagent,
  title={Large Language Model Agent: A Survey on Methodology, Applications and Challenges},
  author={Luo, Junyu and Zhang, Weizhi and Yuan, Ye and Zhao, Yusheng and Yang, Junwei and Gu, Yiyang and others},
  journal={arXiv preprint arXiv:2503.21460},
  year={2025}
}

@article{chen2024llmmas,
  title={A Survey on LLM-based Multi-Agent System: Recent Advances and New Frontiers in Application},
  author={Chen, Shuaihang and Liu, Yuanxing and Han, Wei and Zhang, Weinan and Liu, Ting},
  journal={arXiv preprint arXiv:2412.17481},
  year={2024}
}

@article{wang2023autonomous,
  title={A Survey on Large Language Model based Autonomous Agents},
  author={Wang, Lei and Ma, Chen and Feng, Xue and others},
  journal={arXiv preprint arXiv:2308.11432},
  year={2023}
}

@article{zhang2025agenticrl,
  title={The Landscape of Agentic Reinforcement Learning for LLMs: A Survey},
  author={Zhang, Guibin and Geng, Hejia and Yu, Xiaohang and others},
  journal={arXiv preprint arXiv:2509.02547},
  year={2025}
}

@article{jiang2026sokskills,
  title={SoK: Agentic Skills -- Beyond Tool Use in LLM Agents},
  author={Jiang, Y. and others},
  journal={arXiv preprint arXiv:2602.20867},
  year={2026}
}

@article{xu2026agentskills,
  title={Agent Skills for Large Language Models: Architecture, Acquisition, Security, and the Path Forward},
  author={Xu, Renjun and Yan, Yang},
  journal={arXiv preprint arXiv:2602.12430},
  year={2026}
}

@article{yang2026autoskill,
  title={AutoSkill: Experience-Driven Lifelong Learning via Skill Self-Evolution},
  author={Yang, Yutao and Li, Junsong and Pan, Qianjun and others},
  journal={arXiv preprint arXiv:2603.01145},
  year={2026}
}

@article{zhang2026skillflow,
  title={SkillFlow: Benchmarking Lifelong Skill Discovery and Evolution for Autonomous Agents},
  author={Zhang, Ziao and Shi, Kou and Huang, Shiting and others},
  journal={arXiv preprint arXiv:2604.17308},
  year={2026}
}

@article{wang2023voyager,
  title={Voyager: An Open-Ended Embodied Agent with Large Language Models},
  author={Wang, Guanzhi and Xie, Yuqi and Jiang, Wenhu and others},
  journal={arXiv preprint arXiv:2305.16291},
  year={2023}
}

@article{ouyang2022instructgpt,
  title={Training language models to follow instructions with human feedback},
  author={Ouyang, Long and Wu, Jeffrey and Jiang, Xu and others},
  journal={arXiv preprint arXiv:2203.02155},
  year={2022}
}

@article{deepseek2024llm,
  title={DeepSeek LLM: Scaling Open-Source Language Models with Longtermism},
  author={DeepSeek-AI},
  journal={arXiv preprint arXiv:2401.02954},
  year={2024}
}

@article{dai2024deepseekmoe,
  title={DeepSeekMoE: Towards Ultimate Expert Specialization in Mixture-of-Experts Language Models},
  author={Dai, Damai and others},
  journal={arXiv preprint arXiv:2401.06066},
  year={2024}
}

@article{deepseek2024v2,
  title={DeepSeek-V2: A Strong, Economical, and Efficient Mixture-of-Experts Language Model},
  author={DeepSeek-AI},
  journal={arXiv preprint arXiv:2405.04434},
  year={2024}
}

@article{deepseek2024v3,
  title={DeepSeek-V3 Technical Report},
  author={DeepSeek-AI},
  journal={arXiv preprint arXiv:2412.19437},
  year={2024}
}

@article{deepseek2025r1,
  title={DeepSeek-R1: Incentivizing Reasoning Capability in LLMs via Reinforcement Learning},
  author={DeepSeek-AI},
  journal={arXiv preprint arXiv:2501.12948},
  year={2025}
}

@article{yang2024qwen2,
  title={Qwen2 Technical Report},
  author={Yang, An and others},
  journal={arXiv preprint arXiv:2407.10671},
  year={2024}
}

@article{qwen2024qwen25,
  title={Qwen2.5 Technical Report},
  author={Qwen Team},
  journal={arXiv preprint arXiv:2412.15115},
  year={2024}
}

@article{qwen2025qwen3,
  title={Qwen3 Technical Report},
  author={Qwen Team},
  journal={arXiv preprint arXiv:2505.09388},
  year={2025}
}

  \medskip
  \clearpage

  
\end{document}